%% file: paper.tex
\documentclass{article}
\usepackage{iclr2027_conference,times}
\iclrfinalcopy
\usepackage{amssymb}
\usepackage{xcolor}
\usepackage{graphicx}
\usepackage{subcaption}
\usepackage{booktabs}
\usepackage{multirow}
\usepackage{bm}
\usepackage[most]{tcolorbox}
\usepackage{caption}
\usepackage{placeins}
\usepackage{listings}
\usepackage{float}
\usepackage{enumitem}
\usepackage{hyperref}
\usepackage[noabbrev,nameinlink]{cleveref}

\definecolor{linkcolor}{RGB}{92,92,192}
\hypersetup{
    colorlinks=true,
    linkcolor=linkcolor,
    citecolor=linkcolor,
}

\newcommand{\nm}[1]{#1}
\definecolor{metabg}{HTML}{F1F4F7}
\newfloat{method2}{tbp}{lom}
\floatname{method2}{Method}

\lstdefinestyle{scopepy}{
    basicstyle=\ttfamily\footnotesize,
    keywordstyle=\color{blue!60!black}\bfseries,
    commentstyle=\color{gray!70!black}\itshape,
    stringstyle=\color{purple!60!black},
    language=Python,
    columns=fullflexible,
    keepspaces=true,
    breaklines=true,
    showstringspaces=false,
    frame=single,
    framesep=5pt,
    xleftmargin=6pt,
    aboveskip=4pt,belowskip=2pt,
    morekeywords={def,return,for,in}
}
\newcommand{\yes}{{\color{green!55!black}\checkmark}}
\newcommand{\no}{{\color{red!70!black}\ensuremath{\times}}}
\newcommand{\pt}{{\color{orange!90!black}\ensuremath{\sim}}}

\newcommand{\method}{DaRoPE}  

\newcommand{\prope}{HoPE}                    

\title{RoPE is Dead, Long Live RoPE: Towards \\ Scalable Data-aware Positional Encodings}

\author{
Jarod Lévy$^{1,\ast}$, 
Mathurin Videau$^{1,\ast}$, 
Jad Yehya$^{2}$, 
Jean-Rémi King$^{1}$, \\
\textbf{Stéphane d'Ascoli$^{1,\dagger}$ \& Thomas Moreau$^{2,\dagger}$} \\
\vspace{1.5mm} \\
$^1$Meta AI, Paris \qquad $^2$Inria, Université Paris-Saclay, Palaiseau, France \\
\footnotesize $^\ast$Joint first authors \qquad $^\dagger$Joint last authors
}

\tcbset{
  aibox/.style={
    width=\linewidth,
    top=6pt,
    bottom=3pt,
    colback=blue!6!white,
    colframe=black,
    colbacktitle=black,
    enhanced,
    center,
    attach boxed title to top left={yshift=-0.1in,xshift=0.15in},
    boxed title style={boxrule=0pt,colframe=white,},
  }
}
\newtcolorbox{tkyBox}[2][]{aibox,title=#2,#1}
\definecolor{lightblue}{rgb}{0.22,0.45,0.70}
\newcommand{\paperabstract}{%
Transformers process tokens without any inherent notion of order, making positional encoding a fundamental requirement rather than an architectural refinement. Rotary Position Embedding (RoPE) has become the default positional encoding in modern language models, yet it is heavily biased toward nearby tokens. Existing alternatives have been evaluated under different settings, leaving the literature fragmented and without a clear replacement. We bring structure to this landscape by examining a specific weakness of RoPE: its slow frequency bands, whose wavelengths exceed the training context and expose models to unseen angles during extrapolation. We therefore introduce Data aware RoPE (\method{}), which preserves standard RoPE on the fast bands but replaces absolute position on the slow bands with bounded coordinates learned from contextual representations. Therefore, the slow-band geometry depends on the data rather than only on positional distance. We compare representative encodings under matched conditions across synthetic tasks, symbolic music, genomics, neural signals, and language models spanning 124M to 50B parameters. Across these experiments, \method{} leads on non-text benchmarks, mitigates recency bias, while remaining best or on par in language modeling and length extrapolation. Moreover, the learned coordinates also make the mechanism interpretable, revealing how attention layers leverage contextual information beyond token distance. Together, these results support \method{} as the best overall default among the evaluated methods, when there is no domain-specific reasons to prefer another.
}

\begin{document}
\raggedbottom

\maketitle
\begin{abstract}
\paperabstract
\end{abstract}


\section{Introduction}
\label{sec:intro}

Full-attention has no intrinsic representation of token order, so Transformers typically add a positional encoding (PE)~\citep{vaswani2017attention}. Early models used absolute sinusoidal or learned embeddings~\citep{vaswani2017attention,radford2018gpt,brown2020gpt3}, followed by relative schemes~\citep{shaw2018relative,raffel2020t5,dai2019transformerxl,chi2022kerple,li2024fire}. Rotary Position Embedding (RoPE)~\citep{su2024roformer} is now widely used in modern language models~\citep{chowdhery2023palm,touvron2023llama,jiang2023mistral}. RoPE became popular because it represents relative offsets without learned positional parameters or explicit attention biases, adds overhead linear in sequence length, and remains compatible with FlashAttention~\citep{dao2022flashattention}. These properties have contributed to its widespread adoption across Transformer domains, from language to vision and time-series forecasting~\citep{heo2024ropevit,liu2024timerxl}.

RoPE can, however, induce an average attention decay with relative distance and thereby favor nearby interactions~\citep{su2024roformer,chen2024hope}. Earlier positional encodings were developed and evaluated mainly within comparatively short, fixed contexts. Modern applications such as retrieval-augmented generation, long-document question answering, and extended chains of thought may instead depend on information introduced thousands of tokens earlier. In these settings, distance is a poor proxy for relevance and can cause models to underuse distant evidence in favor of more recent context. In particular, a recent look-alike can override a correct earlier token~\citep{liu2024lostinmiddle,xu2024taskhaystack}. Recent analyses connect this behavior to RoPE's frequency structure, particularly its slow bands, whose wavelengths exceed the training context~\citep{barbero2024round,gopalakrishnan2024pope}. RoPE is also sensitive to the rotary base~\citep{men2024base}, numerical precision~\citep{wang2024bfloat16}, and the long-context training recipe~\citep{gao2025prolong}.

Different domains have adopted different positional conventions, including relative attention in music~\citep{huang2019musictransformer} and absolute embeddings in early genomic models~\citep{ji2021dnabert}. Yet music and genomes both contain motifs that recur at variable distances~\citep{huang2019musictransformer,nguyen2023hyenadna}, making them useful tests of whether proximity is an appropriate prior. Neural time series (EEG) provide a complementary test: its quasi-periodic rhythms recur over time, making proximity a similarly questionable proxy for relevance~\citep{kemp2000sleepedf}.

Despite the broad relevance of positional encoding, evidence on the alternatives to RoPE remains fragmented. Each proposal is evaluated using different model sizes, datasets, context lengths, and metrics, usually against RoPE rather than against one another~\citep{barbero2024round,chen2024hope,gopalakrishnan2024pope,golovneva2024cope,yang2025ropetonope}. Moreover, perplexity,  retrieval, and accuracy in few shot settings measure different capabilities and can rank methods differently~\citep{chen2024hope}. Consequently, it remains unclear which design choices generalize well and scales. RoPE remains the default by inheritance rather than by controlled comparison.

This paper brings structure to the fragmented positional encoding landscape by showing that many methods differ primarily in how they treat RoPE's slow bands, whose wavelengths exceed the training context. This view motivatesa Data aware RoPE (\method{}), which leaves the fast bands unchanged but replaces absolute position on the slow bands with a bounded coordinate learned from contextual token representations. 
%
%
Our contributions are:
\begin{enumerate}[label=\textbf{C\arabic*.}, leftmargin=2.4em]
    \item \textbf{A data-aware positional encoding at RoPE's cost.}
    \method{} repurposes RoPE's slow bands using a per-head content coordinate. It adds a negligible number of parameters, preserves linear overhead and FlashAttention compatibility, and requires no modification at inference.
    
    \item \textbf{A controlled evaluation across domains and scales.}
    Within each experiment, we keep the architecture, data, and training budget fixed, varying only the positional encoding. We evaluate synthetic capabilities, symbolic music, genomics, neural time series, and language modeling across six positional encodings. We benchmark methods at scales up to 1B parameters and further validate the two strongest approaches in a 50B-parameter mixture-of-experts model.
    
    \item \textbf{A practical default.} 
    \method{} combines RoPE-like efficiency, strong language-modeling quality and extrapolation, the strongest overall non-text performance, and greater resistance to recency and retrieval interference. In the absence of domain-specific evidence favoring another encoding, we recommend \method{} as the first positional encoding to evaluate.

\end{enumerate}

\section{From RoPE's slow bands to \method{}}
\label{sec:method}

\subsection{RoPE and its fast and slow bands}
RoPE represents position through rotations at multiple frequencies. Consider a training
sequence of length $L$, with tokens indexed by $m\in\{0,\ldots,L-1\}$. After the query and
key projections, RoPE groups each $d$-dimensional query and key into $d/2$ two-dimensional
bands and rotates band $j$ of token $m$ by
  \begin{equation}
      \tilde q_m^{(j)}
      =
      R\!\left(m\,\theta_j\right)q_m^{(j)},
      \qquad
      R(\phi)
      =
      \begin{pmatrix}
          \cos\phi & -\sin\phi\\
          \sin\phi & \cos\phi
      \end{pmatrix},
      \qquad
      \theta_j=\text{base}^{-2j/d}.
      \label{eq:rope}
  \end{equation}

The base is typically set to $10\text{k}$. The same position-dependent rotation is applied
to the key band $k_m^{(j)}$. Thus, for tokens $m$ and $n$, their contribution to the
query--key score is
\[
\tilde q_m^{(j)\top}\tilde k_n^{(j)}
=q_m^{(j)\top}R\!\big((n-m)\theta_j\big)k_n^{(j)},
\]
because $R(m\theta_j)^{\!\top}R(n\theta_j)=R\big((n-m)\theta_j\big)$. Each band therefore
depends on the relative offset $n-m$, even though its rotation is applied independently to
each token. This per-token operation adds positional overhead linear in $L$ and remains
compatible with FlashAttention.

The bands $j$ operate at different positional scales with wavelength $\lambda_j=2\pi/\theta_j$. We call a band \textbf{fast} if it completes at least one rotation within the training context length $L$, and \textbf{slow} otherwise:
\begin{equation}
  j \ \text{is slow}
  \iff
  \lambda_j>L
  \iff
  \theta_j<2\pi/L,
  \qquad
  K=\#\{j:\theta_j\geq 2\pi/L\}.
  \label{eq:split}
\end{equation}
Thus, bands $j<K$ are fast and bands $j\geq K$ are slow.
Fast bands complete at least one rotation during training, whereas slow bands see only part of a cycle and therefore encounter unseen angles beyond the training context. Prior analyses identify this frequency structure as central to RoPE's behavior~\citep{barbero2024round,chen2024hope}. The rotary base sets the fast--slow boundary: at $L{=}4096$ and $d{=}128$, raising it from $10\text{k}$ to $500\text{k}$ increases the number of slow bands from $18$ to $32$ of $64$.
This base adjustment is a common long-context intervention~\citep{men2024base}.



\begin{table}[t]
    \centering
    \small
    \setlength{\tabcolsep}{5pt}
    \caption{\textbf{Comparison of positional-encoding design properties.}
    ``Data-aware'' indicates that the positional mechanism depends on the input. Throughput
    measures prompt-prefill speed on the 1B architecture at context 4096 on one A100-80GB
    (higher is better), using the median of nine timed iterations after warmup.
    \yes~yes, \pt~partial or with caveat, \no~no.}
    \label{tab:unify}
    \resizebox{\columnwidth}{!}{%
    \begin{tabular}{lccccc}
    \toprule
    Method & Slow-band geometry & Data-aware & Extrapolates & No in-domain & Throughput \\
           &                    & position   & (train-free) & penalty      & (ktokens/s) \\
    \midrule
    RoPE~\citep{su2024roformer}
        & Fixed token index              & \no  & \no  & \yes & \yes ~61.7 \\
    NoPE~\citep{kazemnejad2023nope}
        & None                           & \no  & \pt  & \no  & \yes ~62.4 \\
    HoPE~\citep{chen2024hope}
        & Removed                        & \no  & \yes & \yes & \yes ~62.3 \\
    YaRN~\citep{peng2024yarn}
        & Rescaled token index           & \no  & \yes & \no  & \yes ~62.2 \\
    CoPE~\citep{golovneva2024cope}
        & Pairwise data-aware coordinate & \yes & \pt  & \yes & \no ~~~4.3 \\
    PoPE~\citep{gopalakrishnan2024pope}
        & Token index + phase bias       & \pt  & \pt  & \yes & \pt ~43.4 \\
    \textbf{\method{}}
        & Bounded per-head coordinate    & \yes & \yes & \yes & \yes ~59.8 \\
    \bottomrule
    \end{tabular}
    }
\end{table}

\subsection{Existing approaches and their trade-offs}
With this frequency-based view in place, existing methods can be organized into three broad responses to the slow band problem. Table~\ref{tab:unify} summarizes the different approaches.

\textbf{Extend at inference (YaRN).}
YaRN~\citep{peng2024yarn} rescales RoPE's frequencies at inference to reach longer
contexts. It extrapolates without retraining, but requires the target length and can degrade in-domain perplexity as this introduces a shift between training and inference.

\textbf{Remove the distance prior (NoPE and HoPE).} Decoder-only Transformers
can infer position from the causal mask alone~\citep{haviv2022nope,kazemnejad2023nope,
irie2025petroglyph}. NoPE therefore removes positional rotations entirely.
HoPE~\citep{chen2024hope} instead retains RoPE on the fast bands and sets the slow-band rotation to zero according to Eq.~\ref{eq:split}.

\textbf{Content-dependent and polar alternatives (CoPE and PoPE).}
CoPE~\citep{golovneva2024cope} derives a content-dependent position for each query--key pair by gating their interaction and accumulating those gates between token positions.
This requires pairwise accumulation over the sequence and materializes the full $L\times L$ attention matrix, making CoPE $\mathcal{O}(L^2)$ in memory and inefficient.

PoPE~\citep{gopalakrishnan2024pope} instead disentangles magnitude and phase through a
polar reparameterization. Magnitudes are obtained by applying a softplus to each element,
while the positional phase includes a learned per-component offset:
\begin{equation}
    \langle \tilde q_m, \tilde k_n\rangle = \sum_{c}
    \underbrace{\mathrm{sp}(q_{m,c})\,\mathrm{sp}(k_{n,c})}_{\text{what}}
    \underbrace{\cos\!\big((n{-}m)\,\theta_c + \delta_c\big)}_{\text{where}},
    \label{eq:pope}
\end{equation}
where $\mathrm{sp}$ denotes softplus and $\delta_c\in[-2\pi,0]$ is a learned, input-independent phase bias. PoPE uses
$d$ frequencies $\theta_c$, whereas RoPE uses $d/2$, doubling the width of
$QK^{\!\top}$ and adding an $\mathcal{O}(L^2d)$ term.

The three lines of work disagree on the remedy but converge on the same locus: RoPE's slow~bands. Rescaling them requires knowing the target length. Removing their distance prior avoids unseen rotations, but HoPE gives the freed capacity no new role-, an opportunity it explicitly
notes: ``could be better utilized''~\citep{chen2024hope}. Content-dependent schemes can adapt
position to the input, but existing approaches remain computationally expensive. This suggests a new combination: preserve RoPE's fast-band position clock while placing
an efficient content coordinate on the slow bands.

\subsection{\method{}: Data-aware RoPE}

\method{} is a content-aware positional method that acts only on RoPE's slow bands. It
preserves the index-based fast-band rotations that encode local order, but replaces the token index on each slow band with a bounded, per-head coordinate predicted from the contextual token representation. Slow-band relative phase can therefore reflect which tokens are related rather than only how far apart they occur, while the operation remains a per-token rotary transformation.


\begin{method2}[b]
\caption{\textbf{\method{} pseudocode and slow-band geometry.}
Left: the \method{} rotary code update. Right: fast- and slow-band coordinates under RoPE and \method{}, with the schematic example \emph{the cat is a feline}. RoPE preserves token order, HoPE removes slow-band rotation, and \method{} places \emph{cat} and \emph{feline} nearby in its data-aware coordinate.}
\label{alg:darope}
\begin{minipage}[c]{0.53\linewidth}
\begin{lstlisting}[basicstyle=\ttfamily\scriptsize,
                   backgroundcolor=\color{black!3},
                   frame=single,
                   rulecolor=\color{black!25},
                   columns=fixed,
                   breaklines=false,
                   xleftmargin=3pt,
                   xrightmargin=3pt,
                   aboveskip=1pt,
                   belowskip=0pt]
theta_j = base ** (-2 * arange(d//2) / d)
K = sum(theta_j >= 2*pi/L)

eps = 1e-6
m_bar = clip(m + .5, eps*L, (1-eps)*L)
beta_m = logit(m_bar / L)
c_h = L * sigmoid(w_h.T @ x_m + alpha_h * beta_m)

angle[..., :K] = m * theta_j[:K]
angle[..., K:] = c_h[..., None] * theta_j[K:]
q, k = rotate(q, angle), rotate(k, angle)
\end{lstlisting}
\end{minipage}\hfill
\begin{minipage}[c]{0.44\linewidth}
\centering
\scriptsize
\renewcommand{\arraystretch}{1.25}
\begin{tabular}{lcc}
\toprule
 & Fast bands angle & Slow bands angle \\
\midrule
RoPE & $m\theta$ & $m\theta$ \\
DaRoPE & $m\theta$ & $c_h(x_m)\theta$ \\
\bottomrule
\end{tabular}

\vspace{3pt}
\begin{tabular}{lccccc}
\toprule
Token & the & cat & is & a & feline \\
\midrule
$m$ & 0 & 1 & 2 & 3 & 4 \\
$c_h(x_m)$ & 0.1 & \textbf{1.0} & 3 & 0.7 & \textbf{1.1} \\
\bottomrule
\end{tabular}
\end{minipage}
\end{method2}

Formally, \method{} leaves the fast bands ($j<K$) exactly as RoPE
(Eq.~\ref{eq:rope}). On the slow bands ($j\ge K$), it replaces the absolute position $m$
with a learned per-head content coordinate $c_h(x_m)\in[0,L]$ computed from the token's
hidden state $x_m$:
\begin{align}
    \beta_m   &= \sigma^{-1}\!\left(\tfrac{\bar m}{L}\right)
                 && \text{(fixed positional prior)}, \label{eq:beta}\\
    c_h(x_m)  &= L\cdot\sigma\!\left(w_h^{\!\top} x_m + \alpha_h\,\beta_m\right)
                 && \text{(content coordinate)}, \label{eq:ch}
\end{align}
The formulation uses a learned per-head projection $w_h\in\mathbb{R}^{d_{\text{model}}}$, a learned scalar
$\alpha_h\in\mathbb{R}$, and $\sigma$ the logistic sigmoid bounding the coordinate to
$[0,L]$. We define $\bar m = m+0.5$ and we clip it in code before applying $\sigma^{-1}$. This keeps the positional prior finite outside the training window while the learned coordinate remains bounded. Within the training context, if $w_h{=}0$ and $\alpha_h{=}1$, the two functions cancel and \method{} falls back to RoPE up to a constant offset. The coordinate is data-aware rather than purely content-based: $w_h^\top x_m$ can reorder tokens using context, while $\alpha_h\beta_m$ retains an explicit positional prior whose strength is learned independently by each head. The slow-band rotation then uses $c_h(x_m)$
in place of $m$:
\begin{equation}
    \tilde q_m^{(j)} =
    \begin{cases}
        R\!\left(m\,\theta_j\right)\, q_m^{(j)}, & j<K \ \ \text{(fast: absolute position)},\\[3pt]
        R\!\left(c_h(x_m)\,\theta_j\right)\, q_m^{(j)}, & j\ge K \ \ \text{(slow: content coordinate)},
    \end{cases}
    \label{eq:scope}
\end{equation}
and the same substitution is applied to the keys. By the relative property of $R(\cdot)$,
the slow-band score between $m$ and $n$ depends on $\big(c_h(x_n)-c_h(x_m)\big)\theta_j$:
two tokens with similar content receive similar coordinates and interact as neighbors,
however far apart, while the fast bands keep encoding true position. Because $\sigma$ bounds
$c_h$ to $[0,L]$, slow-band angles stay in their trained range at any evaluation length, so
extrapolation needs no target length. A zero slow-band contribution recovers HoPE. \method{} changes only the rotary angle, adding
$d_{\text{model}}{+}1$ scalars per head, at most $0.3\%$ of model parameters in our evaluated models. The output
remains a per-token rotation of $q,k$, so \method{} keeps RoPE's $\mathcal{O}(n)$ computational cost and
FlashAttention compatibility and requires no target-length-dependent inference-time modification.
The canonical split follows Eq.~\ref{eq:split}. An ablation experiment supports this formulation: removing the positional prior or
making $\alpha$ fixed, per-band, or token-dependent weakens the performance
(Appendix Table~\ref{tab:srope-formulation}).

Method~\ref{alg:darope} summarizes the minimal code update for DaRoPE. Fast bands keep the token index, while slow bands replace it with the data-aware coordinate $c_h(x_m)$, encouraging semantically related tokens to lie close together.

This design is motivated by settings in which relevance is not aligned with linear distance.
For example, the learned slow-band coordinate can create a shortcut between a translated word
and its aligned source word. More broadly, learned word representations can reflect distance in
the syntactic dependency tree rather than linear distance~\citep{hewitt2019structural}, while
neural activity during speech tracks both when information occurs and what it
means~\citep{ding2016cortical,huth2016natural}.

\section{Language Modeling from 124M to 50B}
\label{sec:language-modeling}

In our experiments, we systematically compare RoPE-10k/500k, NoPE, HoPE, PoPE, and
\method{}; CoPE appears only on toy tasks because it does not scale to larger settings
(Table~\ref{tab:unify}).

\begin{figure*}[t]
    \centering
    \includegraphics[width=\textwidth]{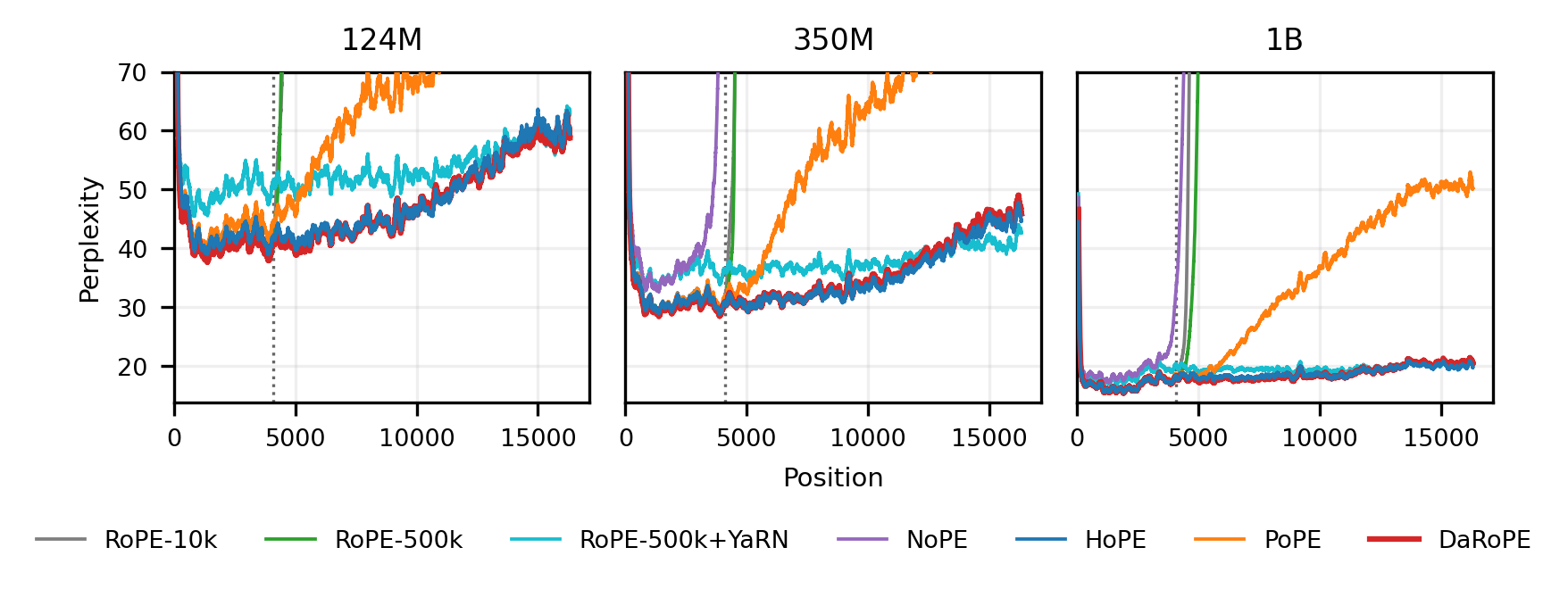}
    \caption{\textbf{Language Models from 124M to 1B: In-Domain and Extrapolation Performance.}
    Per-position perplexity on $N{=}256$ held-out PG-19 documents. Models train at context
    4096 (dotted line) and evaluate to 16k without further training; YaRN is applied to
    RoPE-500k at inference time. Curves use a 128-token moving average; the y-axis is
    clipped at perplexity $70$ for readability.}
    \label{fig:extrap}
\end{figure*}

\paragraph{Dense models.}  Figure~\ref{fig:extrap} reports the perplexity relative to the sequence position for matched decoder-only Transformer at 124M, 350M, and 1B parameters with a 4096 training context, changing only the positional encoding. Using Meta Lingua codebase~\citep{videau2024lingua}, we train the smaller models on FineWeb-Edu~\citep{penedo2024fineweb} and the 1B models on DCLM~\citep{li2024datacomp}. Changing the rotary base is itself a common long-context intervention~\citep{men2024base}, so we evaluate both RoPE-10k and RoPE-500k; the latter also provides a matched-base control for HoPE and \method{}. Architectures, optimization and validation sets are detailed in Appendix~\ref{app:impl:lm}. 

Within the training window, every trained method that retains a positional signal performs
similarly: at 1B over positions 2--4k, these methods cluster at $17.2$--$17.5$ perplexity,
whereas removing position entirely with NoPE reaches $21.0$. YaRN is different because it
rescales frequencies only at inference for the target length; this train--inference shift
degrades its in-window perplexity to $19.1$. Appendix Table~\ref{tab:noreg} reports the
trained-model values on their pretraining distributions.

Extrapolation exposes the slow-band problem directly. Standard RoPE fails once its slow bands reach angles unseen during training; increasing the rotary base moderates this failure but does not solve it. Removing position entirely with NoPE is not a solution, and PoPE also eventually diverges beyond the training window. At 1B over positions 8--16k, HoPE and \method{} remain near $19$ perplexity, while PoPE rises to $42.9$ and RoPE-500k exceeds $450$; the other RoPE and NoPE baselines diverge still further. HoPE and \method{} are therefore the only trained methods whose perplexity remains stable: preserving the fast-band position clock while removing absolute position from the slow bands keeps their geometry in distribution. Both also match the extrapolation of YaRN without target-length-dependent inference-time rescaling. NoPE becomes increasingly competitive as model size grows. However, context extrapolation does not come from simply removing positional encodings: NoPE alone fails beyond the training context window. Instead, as HoPE and \method{} suggest, extrapolation benefits from combining positional encoding with NoPE.

\paragraph{Scaling to a 50B mixture of experts}
HoPE and DaRoPE are the only methods that combine strong in-domain quality and length extrapolation, so we scale each to one matched 49.5B-parameter mixture-of-experts model ($2.57$B active parameters per token). Both train on the same 617B-token mixture at context 4096. More detailed can be found in Appendix~\ref{app:moe-val-loss}.

We first check that scale does not disturb their in-domain capabilities. The models remain remarkably close on 15 in-context benchmarks: DaRoPE averages $54.7$ and HoPE $54.3$. DaRoPE leads on seven tasks, HoPE on eight (Table~\ref{tab:moe-incontext}). This parity spans commonsense reasoning, knowledge, math, and code. The scores are strong for 2.57B active parameters, though differing prompts make prior-work comparisons approximate~\citep{dai2024deepseekmoe, jiang2024mixtral}.

Long-context tasks separate the two models more clearly, and the separation follows a consistent pattern (Table~\ref{tab:moe-longcontext}): \method{}'s advantage concentrates on tasks that require combining evidence spread across a long, natural input. On LongBench (32k) english only subset, multi-document QA alone accounts for about two thirds of \method{}'s 1.6 point average lead (21.1 versus 14.6), while most other categories remain close. Repository-level code completion shows the same effect: on RepoBench, where the relevant code lies in other files of a 32k context, \method{} improves every metric (+5.0 edit similarity, +3.1 exact match, -0.32 answer NLL). Conversely, the gap disappears when long-range aggregation is not required. CrossCodeEval, whose 4k inputs fit inside the training window, is tied, consistent with the in-domain parity above; and on controlled synthetic probes (LongBench synthetic, RULER, BABILong), neither method dominates once task-level variance is taken into account (Appendix \Cref{fig:moe-controlled}). \method{}'s long-context benefit is therefore not a uniform improvement but a specific gain in integrating multi-source evidence beyond the training length, obtained without sacrificing in-domain quality.

\begin{table*}[t]
\centering
\small
\setlength{\tabcolsep}{3pt}
\caption{\textbf{In-context evaluation of the 50B-parameter MoE models.}
Higher is better; bold marks the best accuracy in each column.}
\label{tab:moe-incontext}
\resizebox{\textwidth}{!}{%
\begin{tabular}{@{}l*{16}{c}@{}}
\toprule
&
\multicolumn{8}{c}{Commonsense and reasoning} &
\multicolumn{4}{c}{Knowledge and language} &
\multicolumn{3}{c}{Math and code} &
\multicolumn{1}{c}{Overall} \\
\cmidrule(lr){2-9}
\cmidrule(lr){10-13}
\cmidrule(lr){14-16}
\cmidrule(lr){17-17}

Method
& HSwag & ARC-e & ARC-c & PIQA & OBQA & Wino & CSQA & COPA
& MMLU & RACE & TQA 
& BBH & GSM8K & HEval & MBPP
& Avg. \\
\midrule

\method{}
& 72.9
& \textbf{72.9}
& \textbf{52.1}
& 77.5
& \textbf{43.2}
& \textbf{67.9}
& 55.1
& \textbf{85.0}
& 45.7
& 39.9
& 55.8
& 34.1
& \textbf{29.9}
& \textbf{37.8}
& 50.6
& \textbf{54.7}
\\

 \prope{}
& \textbf{73.6}
& 71.1
& 50.8
& \textbf{78.0}
& 41.6
& 66.1
& \textbf{56.1}
& 82.0
& \textbf{47.2}
& \textbf{40.2}
& \textbf{56.1}
& \textbf{34.3}
& 29.0
& 37.2
& \textbf{51.6}
& 54.3
\\
\bottomrule
\end{tabular}%
}
\end{table*}

\begin{table*}[t]
\centering
\small
\caption{\textbf{Long-context evaluation of the 50B MoE models.} LongBench~\citep{bai2023longbench} uses official metrics averaged within category; CrossCodeEval~\citep{ding2023crosscodeeval} and RepoBench~\citep{liu2023repobench} report edit similarity, exact match, and answer NLL. Higher is better except for NLL; bold marks the best value in each column. Each method is represented by one trained model.}
\label{tab:moe-longcontext}
\resizebox{\textwidth}{!}{%
\begin{tabular}{l*{13}{r}}
\toprule
& \multicolumn{7}{c}{LongBench, english only subset, 32k} & \multicolumn{3}{c}{CrossCodeEval, 4k} & \multicolumn{3}{c}{RepoBench, 32k} \\
\cmidrule(lr){2-8}\cmidrule(lr){9-11}\cmidrule(lr){12-14}
Method & Single QA & Multi QA & Sum. & Synth. & Code & Few-shot & Avg.
& Edit & EM & NLL & Edit & EM & NLL \\
\midrule
\method{} & \textbf{17.4} & \textbf{21.1} & 12.0 & 1.8 & \textbf{57.1} & \textbf{58.4} & \textbf{27.7}
& \textbf{60.3} & \textbf{11.6} & \textbf{1.82} & \textbf{62.6} & \textbf{31.7} & \textbf{2.90} \\
HoPE & 17.0 & 14.6 & \textbf{12.5} & \textbf{3.0} & 56.6 & 55.9 & 26.1
& \textbf{60.3} & 11.4 & 1.83 & 57.6 & 28.6 & 3.22 \\
\bottomrule
\end{tabular}
}
\end{table*}

\paragraph{Efficiency.}
The accuracy gains retain RoPE-like deployment cost (Table~\ref{tab:unify}). At context 4096,
NoPE reaches $62.4$ ktok/s, followed by HoPE at $62.3$, RoPE at $61.7$, and DaRoPE at $59.8$. PoPE falls to $43.4$ because its polar form doubles the query--key width; CoPE reaches only $4.3$. PoPE is $1.4\times$ and CoPE $13.8\times$ slower than DaRoPE before context grows further.

\begin{tkyBox}{HoPE and DaRoPE are the strongest language-modeling defaults}
Both preserve in-domain quality, match YaRN’s extrapolation without inference-time rescaling, and perform comparably in context at 50B. DaRoPE achieves higher average scores on LongBench and RepoBench, particularly when evidence is distributed across long contexts. Both also retain RoPE-like throughput, unlike PoPE and CoPE. 
\end{tkyBox}

\begin{figure*}[!b]
    \centering
    \includegraphics[width=0.95\textwidth]{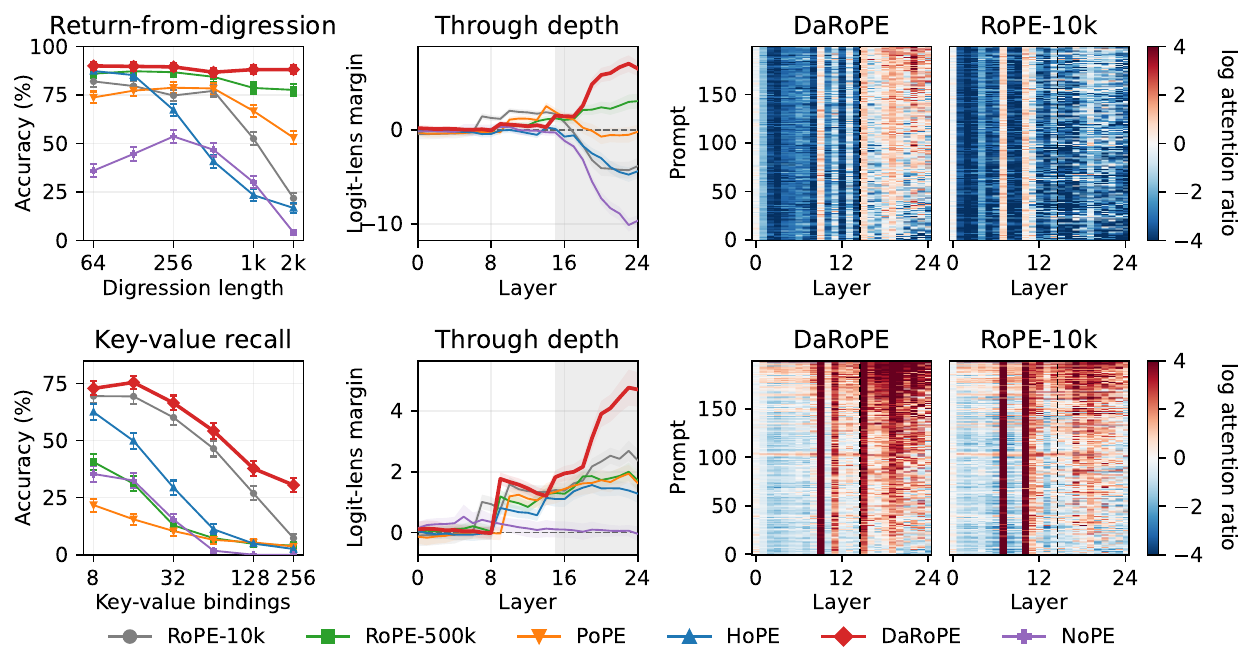}
    \caption{\textbf{Retrieval through model depth.}
    Rows: return-from-digression (top) and key--value recall (bottom). (\emph{left}): accuracy across difficulty ($N{=}800$; 95\% binomial CIs); return compares the correct answer with the recent decoy, while key--value uses exact top-1 selection. (\emph{center}): At the hardest difficulty, correct token's log-probability advantage ($N{=}200$; 95\% CIs; gray marks the final
    ten) for each layer through the final normalization and output head. (\emph{right}): log attention ratio
    $=\log[a(\mathrm{correct})/a(\mathrm{competitor})]$ averaged over all 16 heads; red favors
    correct and blue the recent decoy (digression) or mean of the other $K{-}1$ values (key-value).
    No head or layer is selected.}
    \label{fig:recency-retrieval}
\end{figure*}

\section{DaRoPE Resists Recency and Interference}
\label{sec:recency}
Long sequences rarely follow a single uninterrupted thread. Books return to earlier entities after pages of digression; code reuses symbols across distant blocks; and musical, genomic, and neural motifs recur after variable gaps. In each case, the relevant information may lie far away. A model must be able to retrieve by content rather than proximity.

We test this behavior on the matched 1B models
(Figure~\ref{fig:recency-retrieval}). \emph{Return-from-digression} establishes an early
topic--codeword binding, inserts up to 2048 unrelated tokens, then introduces a plausible but
incorrect recent binding before querying the original topic. The model succeeds only if it
prefers the distant correct codeword over the recent decoy. \emph{Key--value recall} lists
$K$ bindings in random order and queries one randomly positioned key, so distance provides no
clue; increasing $K$ tests retrieval as the number of competing values grows. We report
correct-answer selection at the output in both tasks. Appendix~\ref{app:recency} gives the full
experiment construction.

\method{} is the only method that stays strong throughout both stress tests, improving difficulty-grid average accuracy over the strongest competitor by 5.3 percentage points on \textit{return-from-digression} and 9.6 points on k\textit{ey–value recall} as the number of bindings grows. At the hardest settings, it achieves $88.1\%$ accuracy after a 2048-token digression, compared with $77.6\%$ for RoPE-500k, and $30.5\%$ exact selection with 256 bindings, compared with at most $7.25\%$ for any other method.

The layerwise view explains why. We use a logit lens: after each Transformer block, we apply the model's final normalization and output head to the query representation and measure the NLL margin of the correct answer over its competitor (Appendix~\ref{app:recency}). This diagnostic shows that RoPE-10k recovers the
correct answer internally: its decoded margin reaches $+2.0$ at layer 10. But the signal reverses after layer
16 and finishes at $-3.9$. \method{} instead strengthens the correct answer through the final
layers and ends at $+6.5$. At the final layer, the decoded margin still favors the correct
answer on $86\%$ of \method{} prompts, versus $78\%$ for RoPE-500k, $51\%$ for PoPE, $22\%$
for RoPE-10k, $17\%$ for HoPE, and $3\%$ for NoPE. The attention maps show the
same transition: late-layer attention favors the correct answer on $86\%$ of \method{} prompts and only $1\%$ of RoPE-10k prompts. RoPE can find the answer; \method{} keeps it available until prediction. The pattern is similar for key--value recall. At $K{=}256$, both models retain a positive signal through depth, but \method{} builds a much larger final margin ($+4.71$ versus $+2.37$ on the diagnostic subset). Appendix
Figure~\ref{fig:attention-roster} extends the attention diagnostic to all methods and controls. Randomly reassigning \method{}'s learned coordinates across tokens erases most of both gains, tying them to its data-aware geometry.

These probes isolate a capability that long-form language and motif-heavy sequences demand:
recovering distant content without being overwritten by what came last. The language-modeling and following non-text results show that this advantage survives in real data.

\begin{tkyBox}{\method{} mitigates recency bias and preserves retrieval under interference}
\method{} keeps relevant evidence accessible as distance and interference grow, limiting
late-layer drift toward recent decoys and preserving accuracy through prediction.
\end{tkyBox}

\begin{figure*}[t]
    \centering
    \includegraphics[width=\textwidth]{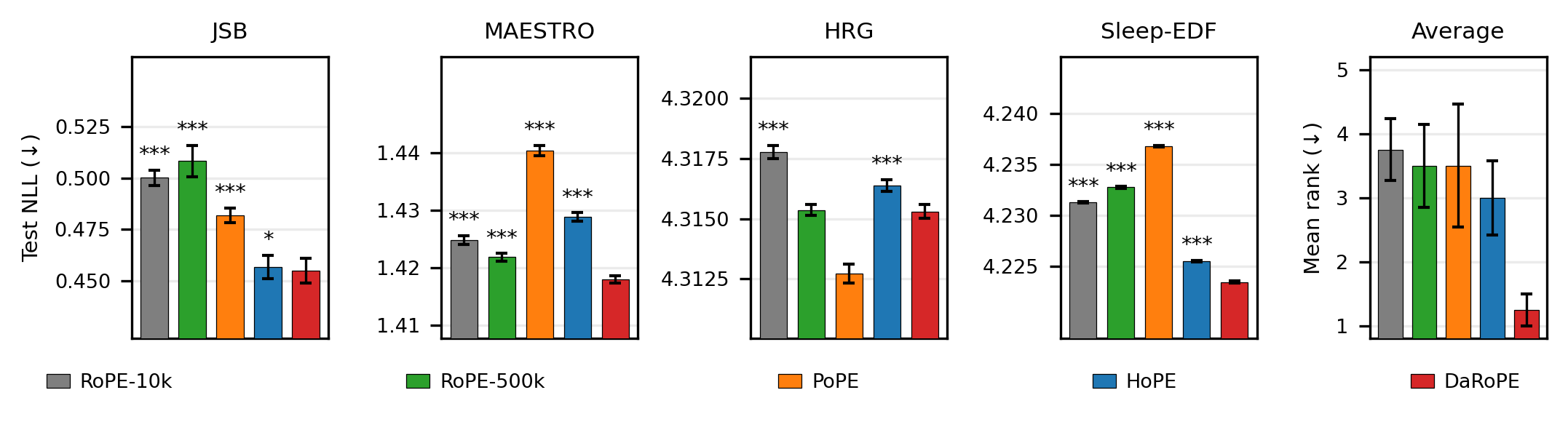}
    \caption{\textbf{Test NLL on four non-textual datasets.} Bars show means across three
    seeds; error bars are 95\% within-example confidence intervals. The right panel reports
    mean rank across the four datasets, with standard errors across datasets. Holm-corrected
    paired tests use test examples across three seeds. Significance is relative to \method{}:
    ${*}\,p<0.05$, ${**}\,p<0.01$, ${***}\,p<0.001$. Test sizes are $N{=}77$ (JSB), $639$ (MAESTRO),
    $8399$ (HRG), and $57{,}782$ (Sleep-EDF). NoPE is off-axis: $1.1634$, $1.5075$,
    $4.3494$, and $4.4320$, respectively.}
    \label{fig:seqmodel}
\end{figure*}

\section{Beyond Language}
\label{sec:nontext}
\paragraph{Music, genomics, and EEG}
 Having established the advantage of HoPE and \method{} for language modeling, we ask whether it extends beyond language. We test the same positional choices on JSB Chorales and MAESTRO symbolic music, the human reference genome (HRG), and Sleep-EDF EEG. Within each domain, every method uses the same architecture, data order, and token budget, with three training seeds. We tune the shared model and optimizer with RoPE-10k. Appendix~\ref{app:nontext} gives preprocessing, architectures, optimization, and evaluation details.

\method{} improves over HoPE on all four datasets: $0.4547$ versus $0.4567$ on JSB,$1.4180$ versus $1.4289$ on MAESTRO, $4.3153$ versus $4.3164$ on HRG, and $4.2235$
versus $4.2255$ on Sleep-EDF (Figure~\ref{fig:seqmodel}). It also systematically beats RoPE-10k. On the two shared music benchmarks, \method{} also improves on the test NLL reported for PoPE in its original paper~\citep{gopalakrishnan2024pope}; the corrected HRG evaluation is discussed in Appendix~\ref{app:impl:genomic}. NoPE confirms that removing position altogether is not enough, while HoPE shows that simply removing the slow-band clock does not always beat RoPE, as on MAESTRO and HRG. PoPE reaches the lowest HRG NLL ($4.3127$), but is much slower: on eight V100s, each PoPE run required $38.9$ hours of training on average, versus $11.8$ hours for \method{}. \method{} leads JSB, MAESTRO, and Sleep-EDF and has the best average rank across all four datasets.

These domains share a useful structure: musical phrases, genomic motifs, and EEG rhythms recur at variable distances. HoPE removes the slow-band distance prior; \method{} goes further and uses
those bands to bring related content together. Its consistent gain over HoPE shows that the learned coordinate transfers beyond text.

\begin{tkyBox}{\method{} is the strongest overall non-text default}
\method{} leads three of four datasets, improves over HoPE on all four, and retains RoPE cost. No other method combines that accuracy and efficiency across non-textual modalities.
\end{tkyBox}

\subsection{Toy tasks}
To isolate the capabilities supported by each positional encoding, we evaluate all methods on a suite of synthetic tasks. The suite includes tasks spanning the language classes of the Chomsky hierarchy, following \citet{deletang2022neural}, as well as five additional tasks designed to probe fine-grained relative representations and data-aware position. Task definitions are given in \Cref{app:tasks}. For each combination of task and positional encoding, we train a four-layer Transformer decoder with a model dimension of 256 on sequences of length 256. For each task, we measure token-level accuracy rather than exact-match accuracy, as exact match can obscure trends by disproportionately penalizing longer outputs. Results use three seeds.

\noindent
\begin{minipage}[t]{0.51\textwidth}
\vspace{0pt}

\Cref{fig:toy-indomain} reports chance-normalized in-domain accuracy averaged across all 15 tasks. CoPE, HoPE, \method{}, and RoPE form the strongest aggregate group, while PoPE and
NoPE trail overall. The aggregate nevertheless hides complementary task profiles: PoPE is strongest on the Chomsky-hierarchy tasks but weaker on retrieval and positional probes, whereas
NoPE particularly struggles when the answer requires position. Per-task results and evaluations
beyond the training length are reported in \Cref{app:toy-cost}.

\end{minipage}
\hfill
\begin{minipage}[t]{0.46\textwidth}
\vspace{0pt}
\centering
\includegraphics[width=0.72\linewidth]{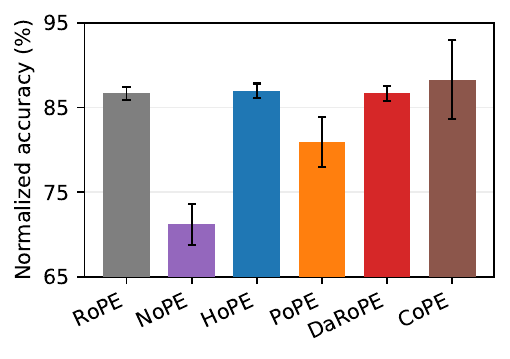}
\captionof{figure}{\textbf{In-domain accuracy.} Mean over 15 tasks; propagated seed SD.}
\label{fig:toy-indomain}
\end{minipage}

\section{Related Work}\label{sec:related}

\paragraph{Positional representations and distance biases.}
Transformers first encoded order with absolute sinusoidal or learned embeddings~\citep{vaswani2017attention,radford2018gpt,brown2020gpt3}. Relative schemes encode pairwise offsets~\citep{shaw2018relative,dai2019transformerxl,raffel2020t5}, while RoPE rotates queries and keys according to token index~\citep{su2024roformer}. ALiBi, KERPLE, and FIRE add distance-dependent biases, whereas xPos adds distance-dependent scaling~\citep{press2022alibi,chi2022kerple,li2024fire,sun2023xpos}. All derive positional geometry from token indices or offsets.

\paragraph{Length extension and rotary frequencies.}
Length-generalization methods broaden training positions through randomization or positional skip-wise training~\citep{ruoss2023randomized,zhu2024pose}, remap coordinates or frequencies as in positional interpolation, YaRN, CLEX, and LongRoPE~\citep{chen2023pi,peng2024yarn,chen2024clex,ding2024longrope}, or target periodic extension directly as in Resonance RoPE and FoPE~\citep{wang2024resonance,hua2025fope}. Extrapolation also depends on the rotary base, numerical precision, and frequency allocation~\citep{men2024base,wang2024bfloat16,barbero2024round}. These approaches improve index-based position; our work asks which bands should encode token index at all.

\paragraph{Removing or learning positional geometry.}
NoPE shows that causal Transformers can infer order without explicit position~\citep{haviv2022nope,kazemnejad2023nope,irie2025petroglyph}; related methods remove RoPE after training or leave selected low-frequency dimensions unrotated~\citep{gelberg2025drope,yang2025ropetonope,barbero2024round,chen2024hope}. Data-aware methods derive distances from content (CoPE), adapt biases (DAPE and GAPE), accumulate transformations (PaTH), or learn input-dependent rotations (CARoPE and Selective RoPE)~\citep{golovneva2024cope,zheng2024dape,ali2026gape,yang2025path,veisi2025carope,movahedi2025selectiverope}. PoPE instead separates content magnitude from positional phase~\citep{gopalakrishnan2024pope}. \method{} targets only the slow bands: they become an interpretable learned reordering, while fast-band RoPE preserves token order at essentially RoPE cost.

\section{Conclusion}\label{sec:conclusion}
RoPE’s slow bands need not remain an absolute clock. \method{} replaces them with bounded, per-head content coordinates while preserving exact fast-band RoPE. Across our evaluations, it matches HoPE on language modeling, retains in-domain quality, extrapolates without inference-time rescaling, runs at near-RoPE cost, leads on non-textual tasks, and reduces recency bias and retrieval interference. Other content axes, band allocations, and long-context fine-tuning remain open. Overall, retain RoPE’s fast positional structure, but let its slow bands adapt to the data.

\subsection*{AI use statement}
Generative AI tools assisted implementation, debugging, data reformatting, figures, code, and
manuscript editing. The authors reviewed all assisted outputs, checked reported results against
the underlying runs and data, and take responsibility for the final content.

\clearpage
\newpage
\bibliographystyle{iclr2027_conference}
\bibliography{paper}

\clearpage
\newpage
\appendix
\raggedbottom
\setlength{\textfloatsep}{10pt plus 2pt minus 2pt}
\setlength{\dbltextfloatsep}{10pt plus 2pt minus 2pt}
\section{Dense Language-Model Details}
\label{app:impl}
This section gives the dense-model architecture, optimization, in-domain evaluation, length
extrapolation protocol, and the formulation ablation referenced in the main paper.

\subsection{Dense-model architecture and evaluation}
\label{app:impl:lm}
All three scales use the Lingua decoder block, a $4096$-token context and the
\texttt{cl100k} tiktoken tokenizer with BOS/EOS. Weights are tied at 124M/350M and untied at 1B. Optimization is AdamW (weight decay $0.1$, gradient clip $1.0$) with a linear warmup followed by cosine decay. Training uses bf16 with \texttt{torch.compile}, TF32
matmuls disabled, and FSDP (\texttt{no\_shard}) across 8$\times$A100-80GB GPUs. Comparisons are made within scale only: 124M/350M train on FineWeb-Edu 10BT~\citep{penedo2024fineweb} and 1B on
DCLM~\citep{li2024datacomp}. The rotary base is $\theta=500\text{k}$ for every method except RoPE-10k and PoPE, which use $\theta=10\text{k}$. Table~\ref{tab:arch} gives the per-scale architecture and optimization.

\begin{table}[t]
\centering
\caption{\textbf{Per-scale language-model architecture and training.} Head dimension is
$d_{\text{model}}/n_{\text{heads}}$; the wavelength split $\theta_j<2\pi/L$ yields $K=d/4$
fast bands at $\theta{=}500\text{k},L{=}4096$.}
\label{tab:arch}
\begin{tabular}{lccc}
\toprule
 & 124M & 350M & 1B \\
\midrule
$d_{\text{model}}$        & 768  & 1024 & 2048 \\
layers                    & 12   & 24   & 25 \\
heads                     & 12   & 16   & 16 \\
head dim $d$              & 64   & 64   & 128 \\
fast bands $K=d/4$        & 16   & 16   & 32 \\
context $L$               & 4096 & 4096 & 4096 \\
weight tying              & yes  & yes  & no \\
peak LR                   & 3e-3 & 2e-3 & 2e-3 \\
LR warmup (steps)         & 2000 & 2000 & 5000 \\
min-LR                    & 1e-6 & 1e-6 & 1e-6 \\
steps                     & 76k  & 152.6k & 152.6k \\
tokens/step               & $\sim$262k & $\sim$131k & $\sim$131k \\
pretraining data          & FineWeb-Edu & FineWeb-Edu & DCLM \\
\bottomrule
\end{tabular}
\end{table}

\paragraph{In-domain perplexity.} Validation perplexity is computed at the training length on a held-out split of the pretraining distribution -- FineWeb-Edu at 124M/350M ($1.94$M tokens) and DCLM at 1B ($4.01$M tokens) -- from the final checkpoint of each run, with every encoding scoring the identical token stream.

\paragraph{Length behaviour.} Figure~\ref{fig:extrap} is computed on PG-19~\citep{rae2019pg19}. We stream the corpus, keep the first $256$ documents with at least $16385$ tokens, truncate each to that length and score the first $16384$ positions in a single forward pass -- no sliding window, no chunking, no fine-tuning and no positional interpolation. The weights are untouched. The value plotted at a position is the mean next-token NLL over the $256$ documents, exponentiated, then smoothed with a $128$-token moving average.
Perplexities quoted in the text average positions $2$--$4$k in window (skipping the first few hundred tokens, where every curve is high) and $8$--$16$k beyond it. YaRN is the single exception to the no-test-time-change rule: following~\citet{peng2024yarn} we apply the NTK-by-parts frequency rescaling to the RoPE-500k checkpoint at inference, with extension factor $s=16384/4096=4$ and use default hyperparameters.

\subsection{\method{} specifics}
\label{app:impl:scope}
The content projection $w_h$ is initialized with standard deviation $0.1$ and the
positional-prior scalar $\alpha_h$ with $0.5$; both are then learned. These two initialization
values were chosen by a small hyperparameter search at the 124M scale and reused unchanged at
350M and 1B. The only added parameters are
$w_h$ ($d_{\text{model}}$) and $\alpha_h$ (1) per head, i.e.
$n_{\text{heads}}\!\cdot\!(d_{\text{model}}{+}1)$ per layer -- under $0.1\%$ of model
parameters at every scale. The fast/slow split uses the wavelength criterion
$\theta_j<2\pi/L$ of \citet{chen2024hope}. For the language models at $\theta{=}500\text{k}$,
$L{=}4096$ this falls at the midpoint (half fast, half slow); the same criterion applied to
the shorter music/genomic contexts moves the split accordingly (a larger slow fraction at
$L{=}1000$). 

\paragraph{Formulation selection.}
We compare five formulations at 124M: the canonical per-head $\alpha_h$, no positional prior,
fixed $\alpha{=}0.5$, per-band $\alpha$, and token-dependent $\alpha$. Each variant uses one
training seed and the final checkpoint; learned content projections are initialized with
standard deviation $0.1$, and learned $\alpha$ terms with $0.5$. Held-out FineWeb-Edu NLL
checks in-domain quality, while mean PG-19 NLL over positions 8--16k selects among formulations
that remain tied in domain. Table~\ref{tab:srope-formulation} is therefore formulation-selection
evidence rather than evaluation on an untouched test set. The canonical per-head scalar gives
the simplest formulation with the strongest selected extrapolation result.

\input{tables/table_formulation_ablation}

\subsection{In-domain validation}
\label{app:noreg}

Repurposing the slow bands does not cost generic quality. Table~\ref{tab:noreg} complements the in-domain and extrapolation results of Figure~\ref{fig:extrap} by listing the exact in-domain
validation perplexity of the full roster at the three dense scales. Every scheme that keeps a positional signal lands within a few tenths of a perplexity point at each scale, with \method{}
matching the strongest baseline; only full NoPE regresses.

\begin{table*}[h]
    \centering
    \begin{tabular}{lccc}
    \toprule
    Method & 124M & 350M & 1B \\
    \midrule
    RoPE-10k & $\nm{18.95}$ & $\nm{15.16}$ & $\nm{17.70}$ \\
    RoPE-500k & $\nm{18.92}$ & $\nm{15.10}$ & $\nm{17.67}$ \\
    NoPE & $\nm{250.63}$ & $\nm{20.15}$ & $\nm{19.56}$ \\
    HoPE & $\nm{18.94}$ & $\nm{15.13}$ & $\nm{17.68}$ \\
    PoPE & $\nm{19.22}$ & $\nm{15.28}$ & $\nm{17.81}$ \\
    \textbf{\method{}} & $\nm{18.95}$ & $\nm{15.09}$ & $\nm{17.70}$ \\
    \bottomrule
    \end{tabular}
    \caption{\textbf{In-domain validation perplexity at the three dense scales.} Perplexity (lower is better) at the training length $L{=}4096$ on a held-out split of each model's own pretraining distribution: FineWeb-Edu at 124M/350M ($1.94$M tokens) and DCLM at 1B ($4.01$M tokens). Values are comparable within a column, not across columns.}
    \label{tab:noreg}
\end{table*}

\section{50B Results and Architecture}\label{app:moe-val-loss}
HoPE and \method{} are compared on 15 in-context benchmarks, per-position PG-19~\citep{rae2019pg19} loss to 16k tokens, and controlled RULER~\citep{hsieh2024ruler}/BABILong~\citep{kuratov2024babilong} tasks. The complete training and architecture configurations follow these results. Each method is represented by one trained model, so these results compare matched checkpoints rather than variability across retraining. The in-context benchmarks are HellaSwag~\citep{zellers2019hellaswag}, ARC~\citep{clark2018arc}, PIQA~\citep{bisk2019piqa}, OpenBookQA~\citep{mihaylov2018openbookqa},
WinoGrande~\citep{sakaguchi2019winogrande}, CommonsenseQA~\citep{talmor2018commonsenseqa}, COPA~\citep{gordon2012copa}, MMLU~\citep{hendrycks2020mmlu}, RACE~\citep{lai2017race}, TriviaQA~\citep{joshi2017triviaqa}, BBH~\citep{suzgun2022bbh}, GSM8K~\citep{cobbe2021gsm8k}, HumanEval~\citep{chen2021humaneval}, and MBPP~\citep{austin2021mbpp}.

\begin{figure}[H]
    \centering
    \includegraphics[width=0.6\linewidth]{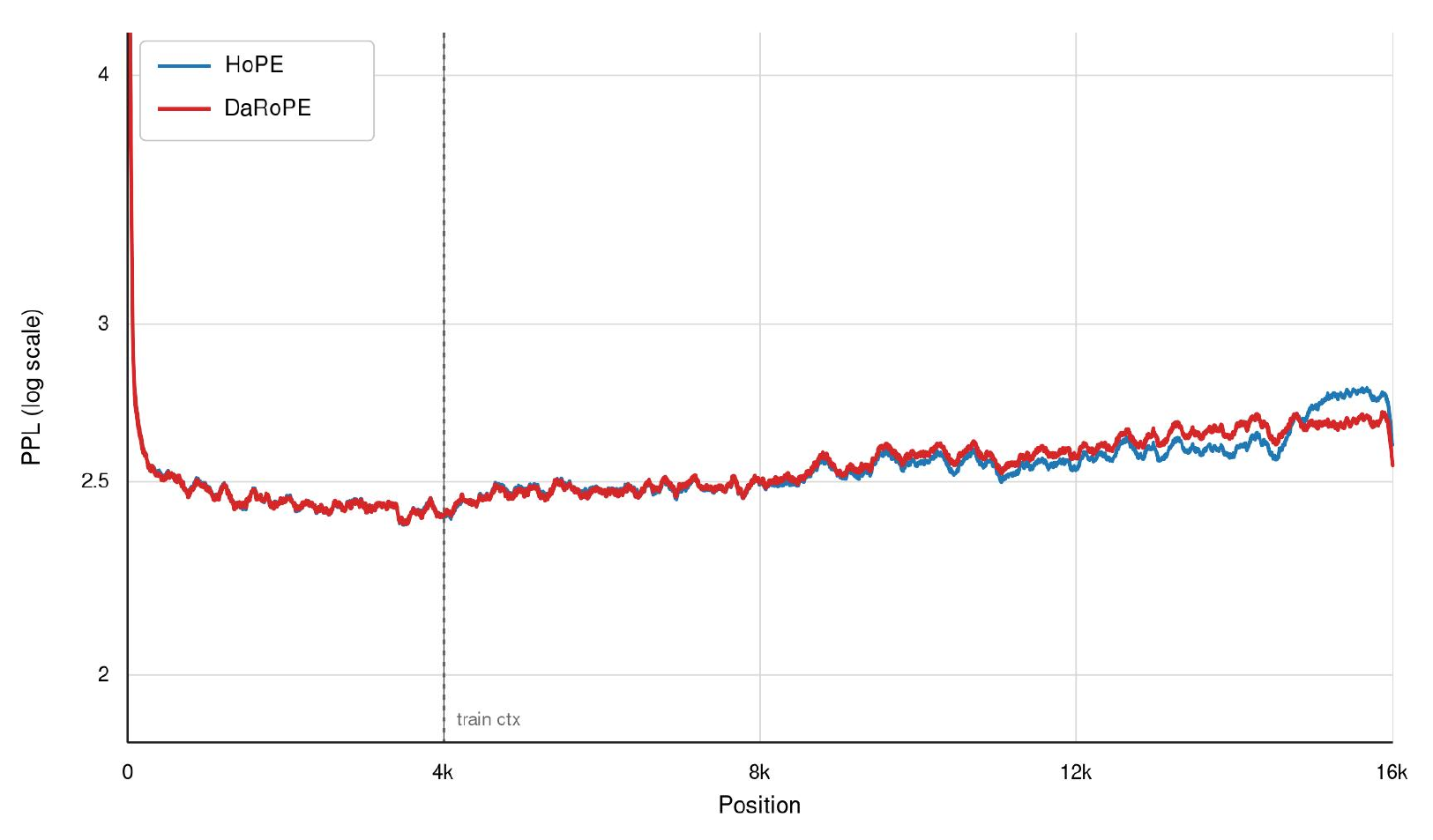}
    \caption{\textbf{
  Per-position validation negative log-likelihood}. The loss is averaged over
  $N=256$ documents for the HoPE and \method{} 50B models. Both models are trained with a context length of 4096 tokens, marked by the
  vertical dashed line, and evaluated on sequences of up to 16,384 tokens. Their curves nearly overlap both within and beyond the
  training window.}
    \label{fig:moe-val-loss}
\end{figure}

The two models remain close throughout PG-19, but their ordering changes with position
(Figure~\ref{fig:moe-val-loss}).
\method{} is marginally lower over the first 8k tokens ($2.4726$ versus $2.4738$ NLL), whereas
HoPE is lower over 8--16k ($2.5913$ versus $2.6054$). Across the full 16k sequence, HoPE
therefore has the lower mean NLL ($2.5325$ versus $2.5390$).

\begin{figure}[H]
    \centering
    \includegraphics[width=0.82\linewidth]{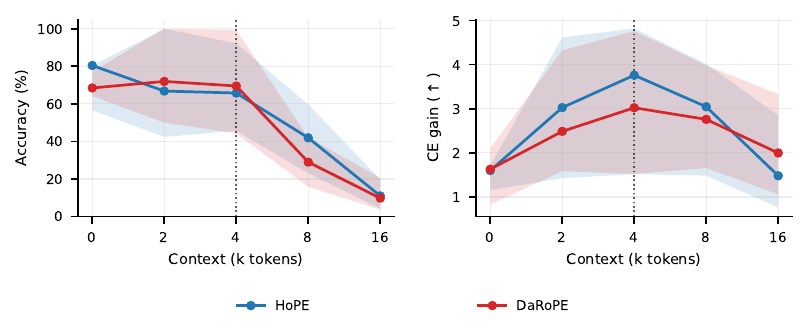}
    \caption{\textbf{Controlled long-context performance across RULER and BABILong.}
    Lines show task-level median accuracy and CE gain; shaded regions show the interquartile range. The dotted line marks the 4k training context. The two methods are mixed across tasks and context lengths.}
    \label{fig:moe-controlled}
\end{figure}

Across RULER and BABILong (Figure~\ref{fig:moe-controlled}), the median advantages reverse
across context lengths and metrics,
while the shaded interquartile ranges overlap broadly. The task-to-task variance is therefore
too large to distinguish the two methods; these controlled evaluations support a tie. 

Note that, for the LongBench results, we excluded news-related data ($<5\%$ of the dataset) to comply with internal policy.The model is evaluated on the english only part as our model is trained on an english only corpora.

\begin{table}[H]
\centering
\small
\caption{Shared training hyperparameters for the HoPE and \method{} MoE models.}
\label{tab:moe-training-hparams}
\setlength{\tabcolsep}{4pt}
\begin{tabular}{@{}ll@{}}
\toprule
\textbf{Training hyperparameter} & \textbf{Value} \\
\midrule
Training steps                    & 65,392 \\
Sequence length                   & 4,096 \\
GPUs                              & 256 \\
Micro-batch / GPU                 & 1 sequence \\
Gradient accumulation             & 9 \\
Global batch size                 & 2,304 sequences \\
Tokens / optimizer step           & 9,437,184 \\
Total training tokens             & 617.1B \\
Training compute                  & $9.52\times 10^{21}$ FLOPs \\
Optimizer                         & AdamW \\
Peak learning rate                & $1.57076\times 10^{-3}$ \\
Minimum learning rate             & $1\times 10^{-6}$ \\
Scheduler                         & Cosine \\
Warmup steps                      & 5,000 \\
Adam $\beta_1,\beta_2$            & $0.9,\,0.95$ \\
Weight decay                      & 0.1 \\
Gradient clipping                 & 0.1 \\
\midrule
\textbf{Data mix}                 & \textbf{Percentage (\%)} \\
\midrule
DCLM                              & 60\% \\
Code                              & 30\% \\
Math                              & 10\% \\
\bottomrule
\end{tabular}
\end{table}

\begin{table}[H]
\centering
\small
\caption{Shared architecture hyperparameters for the MoE models. Fine-grained routed experts with a shared expert follow DeepSeekMoE~\citep{dai2024deepseekmoe}; sigmoid routing with bias-based load balancing follows DeepSeek-V3~\citep{deepseekai2024v3,wang2024auxfree}.}
\label{tab:moe-architecture-hparams}
\setlength{\tabcolsep}{4pt}
\begin{tabular}{@{}ll@{}}
\toprule
\textbf{Architecture hyperparameter} & \textbf{Value} \\
\midrule
Total parameters                    & 49.49B \\
Active parameters / token$^{\dagger}$ & 2.57B \\
Transformer layers                  & 28 \\
Model dimension                     & 2,560 \\
Attention heads                     & 20 \\
Head dimension                      & 128 \\
Vocabulary size                     & 128,256 \\
Maximum training sequence length    & 4,096 \\
Number of routed experts            & 256 \\
Active routed experts / token       & 8 \\
Shared expert                       & Yes \\
Expert FFN hidden dimension         & 864 \\
FFN activation                      & SwiGLU \\
Router score function               & Sigmoid \\
Capacity factor                     & 1.3 \\
Router bias update rate             & 0.005 \\
RoPE base $\theta$                  & 10,000 \\
\bottomrule
\end{tabular}
\end{table}

\FloatBarrier
\section{Non-Textual Sequence Modeling Details}
\label{app:nontext}
The music and genomic models use a patched clone of the PoPE codebase
\citep{gopalakrishnan2024pope}; Sleep-EDF uses the same sanitized training protocol. Within
each domain, all methods share architecture, data order, optimizer, and token budget.

Each domain follows the same simple two-stage recipe. First, we use RoPE-10k to select the architecture and shared training hyperparameters based on validation loss. We then fix the model architecture and training budget across the full model roster, and give \method{} a single small validation-only sweep. The canonical formulation is used for all datasets, except MAESTRO, where we use $16$ slow bands instead of the canonical $17$.

\paragraph{PoPE reproduction and data corrections.}
During reproduction, we identified and reported an upstream finite-data loader issue. We replaced it with a true infinite loader for training and deterministic full-split evaluation for every method. Under this corrected and better-tuned pipeline, PoPE improves on its originally reported JSB and MAESTRO test NLLs~\citep{gopalakrishnan2024pope}, indicating that the published configurations left optimization headroom. HRG additionally uses rebuilt source records that remove duplicated sequence overlap from the test set; its stricter, non-redundant test NLL is therefore higher and not directly comparable with the original reported value.

\subsection{Dataset configurations}
\label{app:impl:music}
The music models use nanoGPT-style decoders with no bias and full-dimension rotation. All four
datasets use the positional-encoding roster and base convention from
Appendix~\ref{app:impl:lm}.

\paragraph{JSB Chorales.}
\textbf{Data:} the Boulanger-Lewandowski chorale corpus~\citep{boulanger2012jsb}, represented
at each timestep by the sounding MIDI notes $21$--$108$, silence, or padding (vocabulary $90$).
\textbf{Model:} $8$ layers, $6$ heads, $d_{\text{model}}{=}192$ ($3.57$M parameters), context
$2048$, dropout $0.3$. \textbf{Training:} AdamW ($\beta_2{=}0.99$, weight decay $0.1$), LR
$6\times10^{-4}$ cosine to $6\times10^{-5}$, $50$ warmup steps, $3000$ iterations, batch $4$.
\textbf{Evaluation:} $N{=}77$ test sequences.

\paragraph{MAESTRO.}
\textbf{Data:} MAESTRO~v3 piano performances~\citep{hawthorne2019maestro}, tokenized with
the REMI representation~\citep{huang2020remi} (pitch $21$--$108$, $32$ velocity bins, $8/4$ beat resolution, chord and tempo
tokens, and PAD/BOS/EOS/MASK). Performances use a $90/5/5$ train/validation/test split; training
uses pitch transposition ($\pm3$ semitones), and sequences are chunked to context $2048$ with
a two-bar overlap. \textbf{Model:} $12$ layers, $12$ heads, $d_{\text{model}}{=}768$
($85.3$M parameters), context $2048$, dropout $0.1$. \textbf{Training:} AdamW
($\beta_2{=}0.99$, weight decay $0.1$), LR $2.15\times10^{-4}$ cosine to
$2.15\times10^{-5}$, $1081$ warmup steps, $54{,}045$ iterations, effective batch $4$
($426.6$M token presentations). \textbf{Evaluation:} $N{=}639$ test sequences.

\paragraph{Human reference genome.}
\label{app:impl:genomic}
\textbf{Data:} the human reference genome (HRG) dataset of \citet{dalla-torre2025nucleotide}, built from GRCh38~\citep{schneider2017grch38}, uppercased with non-ACGT bases mapped to
\texttt{N} and tokenized into $6$-mers (vocabulary $4107$). Chromosomes are split into train
(chr1--20, X, Y), validation (chr21), and test (chr22). We remove duplicated overlap from the
$6200$-base source records and construct $6100$-base examples with $50$-base overlap
(stride $6050$), adding a $0$--$99$-base start jitter during training only.
\textbf{Model:} $306.25$M parameters, $24$ layers, $16$ heads,
$d_{\text{model}}{=}1024$, context $1000$, no dropout, full-dimension rotation.
\textbf{Training:} AdamW ($\beta_2{=}0.999$, weight decay $10^{-2}$), LR
$3.7\times10^{-4}$ cosine to $2.5\times10^{-5}$, $3837$ warmup steps, $95{,}921$
iterations, effective batch $64$ across $8$ GPUs ($63{,}936$ tokens/step, $6.13$B tokens).
\textbf{Evaluation:} held-out test NLL on $N{=}8399$ sequences.

\paragraph{Sleep-EDF.}
\label{app:impl:sleep}
\textbf{Data:} the Sleep Cassette subset of Sleep-EDF Expanded~\citep{kemp2000sleepedf} from PhysioNet~\citep{pollard2026physionet},
comprising $153$ overnight recordings from $78$ subjects, with Fpz--Cz EEG at $100$ Hz
(Sleep Telemetry excluded). Each night is trimmed from the first to last scored sleep stage
plus $30$ minutes on each side, normalized by its median and interquartile range, and clipped
to $\pm20$. The age-stratified, subject-disjoint train/validation/test split contains
$62/8/8$ subjects ($122/15/16$ nights) and $464.9$M/$63.2$M/$59.2$M tokens. Each sample is
mapped through $255$ training-set quantile boundaries to a $256$-token vocabulary; no patching,
vector quantization, continuous head, or sleep-stage labels are used.
\textbf{Model:} $4.79$M parameters, $6$ pre-norm layers, $8$ heads,
$d_{\text{model}}{=}256$, a $4\times$ GELU MLP, RMSNorm~\citep{zhang2019rmsnorm}, no bias or dropout, tied embeddings,
full-dimension rotation, context $1024$. \textbf{Training:} AdamW
($\beta{=}0.9/0.99$, weight decay $10^{-2}$, gradient clip $1.0$), LR
$6\times10^{-4}$ cosine to $6\times10^{-5}$ after $200$ warmup steps, batch $16$, and
$20{,}000$ updates for each of three seeds. \textbf{Evaluation:} deterministic non-overlapping
windows over $N{=}57{,}782$ test sequences ($59.2$M tokens).

\subsection{Test scoring and paired significance}
Comparisons are paired: for a baseline $b$ we test the per-sequence differences
$d_i=\mathrm{NLL}_i(b)-\mathrm{NLL}_i(\method{})$, after first averaging each test example across the three seeds. We apply two-sided paired $t$-tests and Holm correction across the four baseline comparisons within each domain. Figure~\ref{fig:seqmodel} displays a significance marker when the corrected test is significant and \method{} has lower NLL. Error bars are $95\%$ within-example intervals across the paired methods.

\begin{figure*}[b!]
    \centering
    \includegraphics[width=\textwidth]{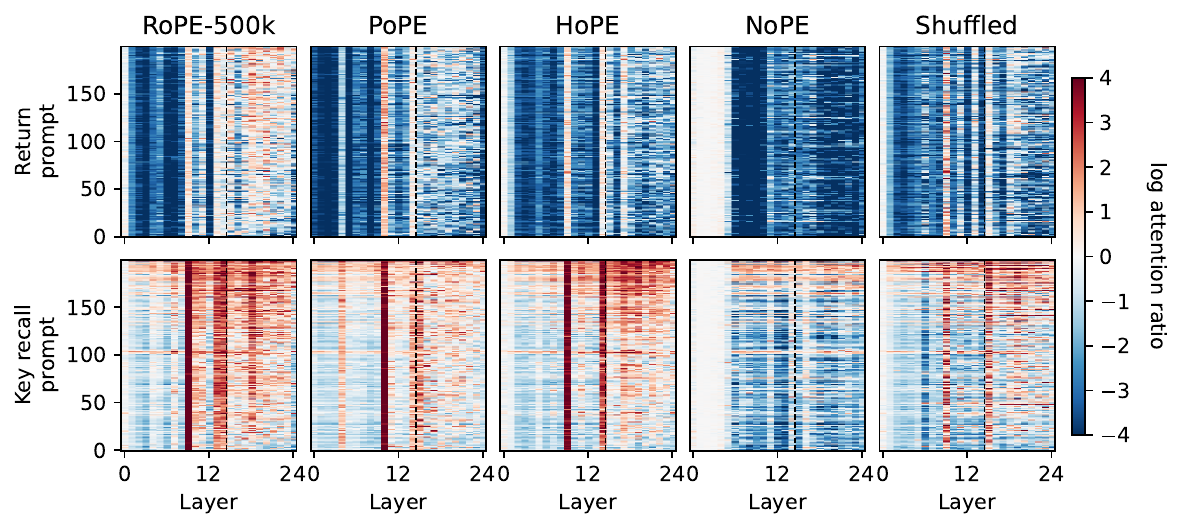}
    \caption{\textbf{Attention ratios for the remaining methods and coordinate intervention.}
    Rows show return-from-digression at $D{=}2048$ (top) and key--value recall at $K{=}256$
    (bottom), matching Figure~\ref{fig:recency-retrieval}. Columns show RoPE-500k, PoPE,
    HoPE, NoPE, and the trained \method{} model after its coordinates are shuffled across token
    positions. Each cell averages all 16 heads in that layer. Red favors the correct value;
    blue favors the recent decoy or mean competitor. All panels use the same $N{=}200$
    prompts, row order, and color scale; the dashed line precedes layers 15--24.}
    \label{fig:attention-roster}
\end{figure*}

\section{Recency and Content Coordinates}

\label{app:recency}
\textbf{Models and instances.} All numbers come from the 1B roster, evaluated
at the $4096$-token training context. Instances are generated from a fixed seed independently
of the model, so the six positional schemes are scored on byte-identical prompts and all
comparisons are paired: $N{=}800$ instances per point and
$D \in \{64,128,256,512,1024,2048\}$ with per-instance jitter $\pm15\%$.
Key--value recall uses $K \in \{8,16,32,64,128,256\}$ and $800$ instances per point.

\textbf{Return-from-digression.} An antecedent binds an answer to a topic; an off-topic
digression padded to $D$ tokens follows; a decoy binds a different answer to a second topic at
the end of the digression; the continuation restates the first topic only. The scored token is
the answer, and the margin is NLL(decoy) $-$ NLL(correct):
\noindent\begin{minipage}{\columnwidth}
\centering
\small
\begin{tabular}{@{}p{0.72\columnwidth}l@{}}
\texttt{[off-topic filler ...]} & \\
\texttt{Zaryndor417 report: codeword = quartz.} & \emph{antecedent} \\
\texttt{[off-topic filler, padded to $D$ tokens ...]} & \\
\texttt{Mournhold263 report: codeword = lantern.} & \emph{recent decoy} \\
\texttt{Summary of Zaryndor417: codeword =} & \emph{continuation}
\end{tabular}
\end{minipage}

\textbf{Key--value recall.} $K$ key--value bindings are listed in random order and one key is
queried again. Its binding sits at a random rank, so distance carries no information. The
margin is the mean NLL of the $K{-}1$ competing values minus the NLL of the correct one;
exact selection requires the correct value to rank first.

\textbf{Layerwise diagnostics.} The mechanism panels use the hardest displayed settings,
$D{=}2048$ and $K{=}256$, with $N{=}200$ model-independent prompts. After block $\ell$, let
$h_\ell$ be the hidden state at the final query position and $W_v$ the output vector for token
$v$. The logit-lens margin is
\[
M_\ell =
\left\langle \operatorname{Norm}(h_\ell),
W_{\mathrm{correct}}-\overline{W}_{\mathrm{competitors}}\right\rangle ,
\]
where the bar is the recent decoy for return-from-digression and the mean of the $K{-}1$
competing values for key--value recall. We apply the model's final normalization and output
head at every layer. Thus $M_\ell$ asks which answer is decodable at that depth; only the final layer is the model's actual output. 

For attention, we first average the final query's attention probability across all 16 heads
within a layer, then report the log ratio between the correct value and the recent decoy or
mean competing value. Positive values favor the correct binding. No layer or head is selected.
Prompt rows are sorted once by \method{}'s mean ratio over layers 15--24 and reused unchanged for
every method.

The near-white first layers for NoPE do not indicate missing attention. White means a log ratio
near zero: correct and competing values receive similar attention. Without an explicit
positional signal, early NoPE layers have not yet separated candidates with the same local format.
Later contextual representations can still become order-sensitive through the causal network,
which explains why a strong preference emerges only after several layers.

\textbf{Coordinate intervention.} We keep \method{}'s trained weights and the complete set of
coordinates fixed, but randomly reassign those coordinates across token positions at
inference. This preserves their values while breaking the token--coordinate correspondence.
At the final layer, shuffling reduces the paired margin by $8.79\pm0.77$ nats on
return-from-digression and $3.76\pm0.47$ on key--value recall (95\% CIs), favoring native
coordinates on $96\%$ and $93\%$ of prompts.

\begin{figure}[t!]
    \centering
        \includegraphics[width=\textwidth]{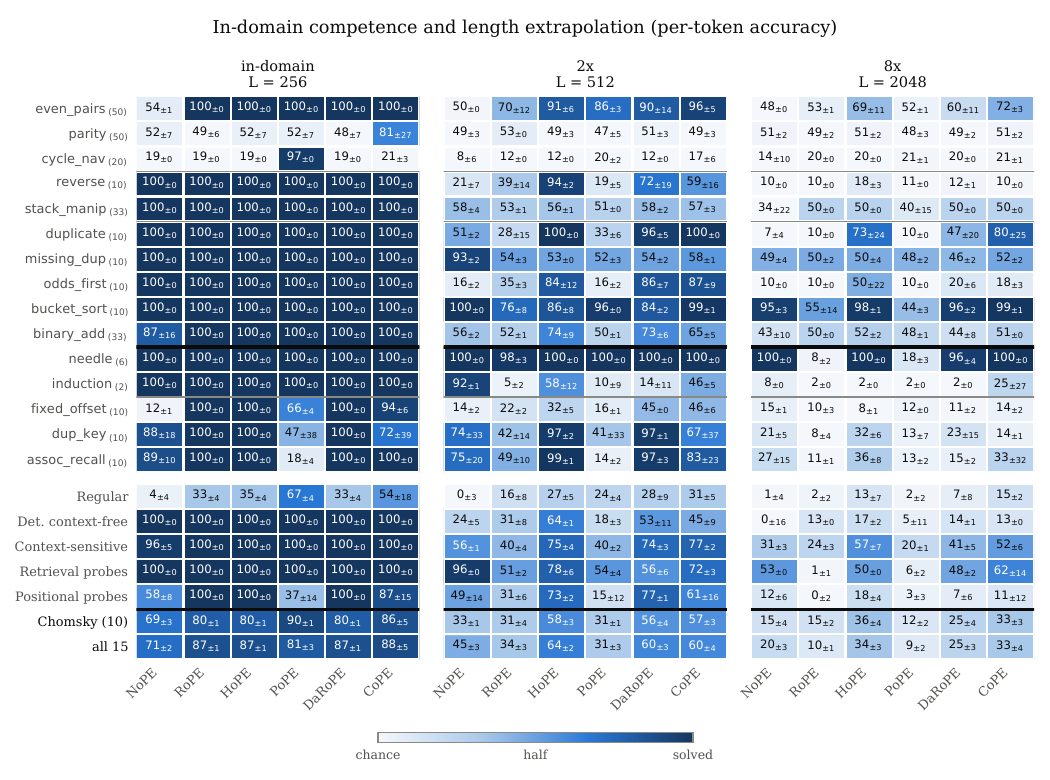}
    \caption{\textbf{Full synthetic-task results.} Per-task token accuracy (mean $\pm$ SD across seeds) in domain and at $2\times$ and $8\times$ the training length, with task-family and aggregate summaries.}
    \label{fig:toy-full-table}
\end{figure}

\section{Synthetic Tasks}\label{app:toy-cost}

\subsection{Task descriptions}
\label{app:tasks}

We evaluate on 15 synthetic tasks in three groups. All are rendered as ASCII strings with a
task-specific prefix, tokenised at byte level, and supervised only on the output span. ``Answer''
below distinguishes a \emph{single} output symbol from a \emph{sequence}; this matters for scoring,
because per-token accuracy and exact-match coincide on single-symbol tasks, whereas on sequence
tasks exact-match requires every symbol to be right and its chance level collapses to $\approx 0$.
Chance is $1/|\mathcal{A}|$ for the answer alphabet $\mathcal{A}$.

\subsection{Chomsky-hierarchy benchmark (15 tasks)}

The Chomsky-hierarchy benchmark \citep{deletang2022neural} is grouped by the level of the formal-language hierarchy required to solve it. Ten are learnable by a 4-layer transformer and carry the main results;
\texttt{modular\_arithmetic}, \texttt{modular\_arithmetic\_brackets} and \texttt{solve\_equation} are at chance for every positional encoding at that depth; \texttt{binary\_multiplication} and \texttt{compute\_sqrt} reach exact-match $0$ at depth 4 with a 256-token context and are excluded from our results.

\begin{table}[H]
\centering\small
\begin{tabular}{@{}llllp{0.31\linewidth}@{}}
\toprule
Task & Example & Answer & $\langle$ch$\rangle$ & Description \\
\midrule
\multicolumn{5}{@{}l}{\textit{Regular}} \\
\texttt{even\_pairs}       & \texttt{EP:01110011} $\to$ \texttt{1}        & single & 50 & Is the number of adjacent differing pairs even? \\
\texttt{parity}            & \texttt{P:01110011} $\to$ \texttt{1}         & single & 50 & Parity of the number of \texttt{1}s. \\
\texttt{cycle\_navigation} & \texttt{CY:11220022} $\to$ \texttt{2}        & single & 20 & Position on a 5-cycle after a walk (\texttt{0}/\texttt{1}/\texttt{2} = left/stay/right). \\
\texttt{modular\_arithmetic} & \texttt{MA:2-3*0+4} $\to$ \texttt{1}       & single & 20 & Evaluate a flat expression mod 5 (the paper's \emph{simple} variant; no brackets). \\
\addlinespace
\multicolumn{5}{@{}l}{\textit{Deterministic context-free}} \\
\texttt{reverse\_string}   & \texttt{R:45790189} $\to$ \texttt{98109754}  & sequence & 10 & Reverse the input. \\
\texttt{stack\_manipulation} & \texttt{ST:11102442} $\to$ \texttt{1111}   & sequence & 33 & Run a stack program (\texttt{2}=pop, \texttt{3}/\texttt{4}=push) and emit the final stack. \\
\texttt{modular\_arithmetic\_brackets} & \texttt{MB:(-2+(3))} $\to$ \texttt{1} & single & 20 & Evaluate a \emph{bracketed} expression mod 5; needs a nesting stack. \\
\texttt{solve\_equation}   & \texttt{EQ:(2+-x)=4} $\to$ \texttt{3}        & single & 20 & Solve for $x$ mod 5; requires inverting the expression. \\
\addlinespace
\multicolumn{5}{@{}l}{\textit{Context-sensitive}} \\
\texttt{duplicate\_string} & \texttt{D:45790189} $\to$ \texttt{4579\dots0189} & sequence & 10 & Emit the input twice ($ss$). \\
\texttt{missing\_duplicate\_string} & \texttt{MD:21110111} $\to$ \texttt{0}   & single & 10 & One symbol of a duplicated string $ss$ is masked; recover it. \\
\texttt{odds\_first}       & \texttt{OF:45790189} $\to$ \texttt{59194708} & sequence & 10 & Emit odd-indexed symbols, then even-indexed ones. \\
\texttt{binary\_addition}  & \texttt{BA:111+0111} $\to$ \texttt{10101}    & sequence & 33 & Add two little-endian binary numbers. \\
\texttt{bucket\_sort}      & \texttt{S:45790189} $\to$ \texttt{01457899}  & sequence & 10 & Sort the symbols of a fixed alphabet. \\
\texttt{binary\_multiplication}$^{\times}$ & \texttt{BM:111*0111} $\to$ \texttt{0100011} & sequence & $\approx$50 & Multiply two little-endian binary numbers. \\
\texttt{compute\_sqrt}$^{\times}$ & \texttt{SQ:11110011} $\to$ \texttt{1111} & sequence & $\approx$50 & $\lfloor\sqrt{n}\rfloor$ of a big-endian binary number. \\
\bottomrule
\end{tabular}
\caption{Chomsky-hierarchy tasks. $^{\times}$ excluded from our results: at depth 4 with a 256-token training context, exact-match is $0$ for every encoding we ran them with, so no instance is solved
\emph{in this configuration}.}
\end{table}

\subsection{Retrieval probes (2 tasks)}

Two retrieval tasks that ship with our codebase but are \emph{not} part of the benchmark. We keep
them separate because they behave very differently from the benchmark tasks: they are the only place
where removing positional information (NoPE) is a large win, and pooling them with the benchmark
inflates NoPE's average.

\begin{table}[H]
\centering\small
\begin{tabular}{@{}llllp{0.50\linewidth}@{}}
\toprule
Task & Example & Answer & $\langle$ch$\rangle$ & Description \\
\midrule
\texttt{needle}    & \texttt{N:45790189} $\to$ \texttt{00000001} & sequence & 6 & At each step, has the current symbol occurred earlier? \\
\texttt{induction} & \texttt{I:55791289} $\to$ \texttt{05000001} & sequence & 2 & Given \dots\,\texttt{AB}\,\dots\,\texttt{A}, predict \texttt{B} (induction head). \\
\bottomrule
\end{tabular}
\caption{Retrieval probes. Pure content matching: the answer never depends on absolute position.}
\end{table}

\subsection{Positional probes (3 tasks)}

Three tasks we introduce to isolate the positional axis. They share their rendering, alphabets and
answer format exactly, and differ \emph{only} in how much positional information the answer
requires.

\begin{table}[H]
\centering\small
\begin{tabular}{@{}llllp{0.44\linewidth}@{}}
\toprule
Task & Example & Answer & $\langle$ch$\rangle$ & Description \\
\midrule
\texttt{fixed\_offset}    & \texttt{FO:3141592\#0003} $\to$ \texttt{5} & single & 10 & \textbf{Pure position:} return the digit $k$ places from the end. \\
\texttt{dup\_key\_recall} & \texttt{DK:a3b7a9c1?a} $\to$ \texttt{9}    & single & 10 & \textbf{Position\,+\,content:} the query key occurs twice; return the \emph{later} value. \\
\texttt{assoc\_recall}    & \texttt{AR:a3b7c1?b} $\to$ \texttt{7}      & single & 10 & \textbf{Content only:} the query key occurs once; return its value. \\
\bottomrule
\end{tabular}
\caption{Positional probes. \texttt{assoc\_recall} is the control: it is identical to
\texttt{dup\_key\_recall} except that the query key is unique, so any gap between the two is
attributable to the positional requirement rather than to capacity.}
\end{table}

\end{document}

%% file: tables/table_formulation_ablation.tex
\begin{table}[t]
\centering
\begin{tabular}{lcc}
\toprule
Variant & In-domain NLL $\downarrow$ & 8--16k NLL $\downarrow$ \\
\midrule
\textbf{\method{}} (per-head $\alpha$) & 2.9447 & 3.957 \\
No $\beta$ & 2.9420 & 4.066 \\
Fixed $\alpha$ & 2.9437 & 4.449 \\
Per-band $\alpha$ & 2.9439 & 4.815 \\
Dynamic $\alpha$ & 2.9451 & 4.281 \\
\bottomrule
\end{tabular}
\caption{\textbf{DaRoPE formulation ablation at 124M.} One training seed; held-out in-domain and 8--16k NLL, lower is better.}
\label{tab:srope-formulation}
\end{table}